\documentclass[letterpaper, 10 pt, conference]{ieeeconf}  

\IEEEoverridecommandlockouts                              

\usepackage[noadjust]{cite}

\usepackage{graphicx}
\graphicspath{{./graphics/}}
\usepackage[utf8]{inputenc}
\usepackage{multirow}

\usepackage{todonotes}
\usepackage{xcolor}

\usepackage[printwatermark]{xwatermark}
\usepackage{tikz}

\usepackage{amsmath}
\title{\LARGE \bf A Fleet Learning Architecture for Enhanced Behavior Predictions during Challenging External Conditions}

\author{Florian~Wirthmüller\textsuperscript{\includegraphics[scale=0.4]{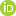}},
	Marvin~Klimke\textsuperscript{\includegraphics[scale=0.4]{orcidlogo.png}},
        Julian~Schlechtriemen\textsuperscript{\includegraphics[scale=0.4]{orcidlogo.png}},
        Jochen~Hipp\textsuperscript{\includegraphics[scale=0.4]{orcidlogo.png}}
        and~Manfred~Reichert\textsuperscript{\includegraphics[scale=0.4]{orcidlogo.png}}%
\thanks{F. Wirthmüller, M. Klimke, J. Schlechtriemen and J. Hipp are with Mercedes-Benz AG, Böblingen, Germany, \newline E-Mail: \{first\_name.last\_name\}@daimler.com}
\thanks{F. Wirthmüller and M. Reichert are with the Institute of Databases and Information Systems (DBIS), Ulm University, Ulm, Germany,\newline E-Mail: \{first\_name.last\_name\}@uni-ulm.de}%
\thanks{J. Schlechtriemen is with the Institute of Realtime Learning Systems at the University of Siegen, Siegen, Germany}
\thanks{ORCID (ordered as authors above): \newline \href{https://orcid.org/0000-0002-9732-2561}{https://orcid.org/0000-0002-9732-2561};\newline \href{https://orcid.org/0000-0003-2647-9673}{https://orcid.org/0000-0003-2647-9673};\newline \href{https://orcid.org/0000-0002-9130-061X}{https://orcid.org/0000-0002-9130-061X};\newline \href{https://orcid.org/0000-0002-9037-9899}{https://orcid.org/0000-0002-9037-9899};\newline \href{https://orcid.org/0000-0003-2536-4153}{https://orcid.org/0000-0003-2536-4153}}\thanks{\copyright~2020 IEEE. Personal use of this material is permitted. Permission from IEEE must be obtained for all other uses, in any current or future media, including reprinting/republishing this material for advertising or promotional purposes, creating new collective works, for resale or redistribution to servers or lists, or reuse of any copyrighted component of this work in other works.}
}

\usepackage{url}
\usepackage{hyperref}

\let\labelindent\relax
\usepackage{enumitem}

\begin{document}

\IEEEoverridecommandlockouts
\pubid{\copyright~2020 IEEE}

\maketitle
\pagestyle{empty}


\begin{abstract}

Already today, driver assistance systems help to make daily traffic more comfortable and safer. However, there are still situations that are quite rare but are hard to handle at the same time. In order to cope with these situations and to bridge the gap towards fully automated driving, it becomes necessary to not only collect enormous amounts of data but rather the right ones. This data can be used to develop and validate the systems through machine learning and simulation pipelines. 
Along this line this paper presents a fleet learning-based architecture that enables continuous improvements of systems predicting the movement of surrounding traffic participants. Moreover, the presented architecture is applied to a testing vehicle in order to prove the fundamental feasibility of the system. Finally, it is shown that the system collects meaningful data which are helpful to improve the underlying prediction systems.
\end{abstract}

\section{Introduction}

\begin{figure}[t!]
\centering\includegraphics[width=0.48\textwidth]{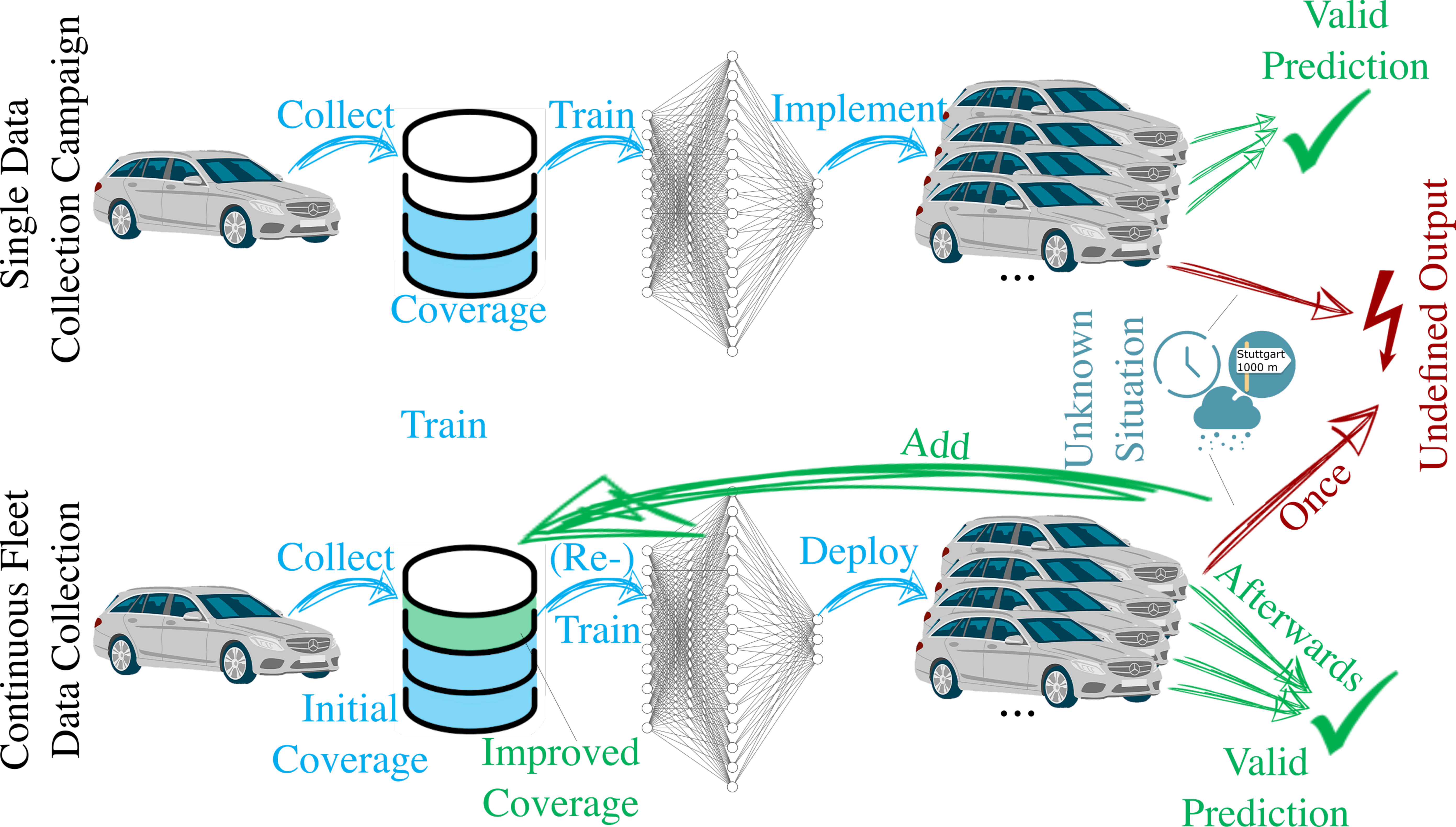}\caption{Current software modules for predicting the movement of surrounding traffic participants are developed through an unidirectional process with a single data collection campaign and model training. This is shown in the upper part of the illustration. In this context a motion prediction module can be thought of as any machine learning model, beeing illustrated as simple neural network. The lower part of the image shows our idea of a continuous fleet data collection in an illustrative way.}\label{fig:intro}
\end{figure}

Driver assistance systems are on the rise and help to prevent accidents and to support drivers in various ways more and more frequently. Thereby, modules predicting future motions of surrounding traffic participants constitute a central piece of such system's intelligence. As shown in \cite{wirthmueller2020}, it is beneficial to integrate external information such as knowledge about weather or traffic conditions into these prediction modules. Therefore, the systems are enabled to deal with rarely occuring and nevertheless challenging conditions, resulting in increased system performances as well as benefits for the drivers in general and especially during challenging conditions. As a prerequisite for developing such context-aware motion prediction modules huge amounts of data need to be collected. But it is not only about gathering the pure amount of data but rather about collecting the right data. This means to collect data during conditions where current systems face problems and data which facilitates developers to improve their systems. As at least some of these conditions occur rather rarely, it is crucial to collect corresponding data with a large fleet of vehicles to ensure a good coverage of all kinds of situations. Hence, this work presents a data collection architecture enabling continuously improving motion predictions over time. Besides, we demonstrate the fundamental feasibility of the approach by integrating it into a testing vehicle. \autoref{fig:intro} illustrates the idea of such a fleet learning architecture indicating its potential for system improvements.

The remainder of this work is structured as follows: \autoref{sec:rel_work} gives an overview of related works. \autoref{sec:arch} introduces the new fleet learning-based architecture concept. The concept enables the detection of challenging conditions with respect to behavior prediction. In particular, it allows
re-parametrizing onboard prediction modules in order to achieve improved prediction performances. Afterwards, \autoref{sec:mem} and \autoref{sec:application} describe the development of the desired prediction watchdog and the prototypical realization of the needed onboard components in a testing vehicle. \autoref{sec:conclusion} summarizes and concludes the article.

\pubidadjcol

\section{Related Work}\label{sec:rel_work}

For the present study, works dealing with the development of systems intended for data collection and management (\autoref{subsec:rel_work_collection}) as well as such dealing with behavior prediction (\autoref{subsec:rel_work_prediction}) are of particular interest. After introducing and categorizing characteristic approaches, \autoref{subsec:rel_work_discussion} discusses them and deduces the contribution of this article.


\subsection{Data Collection and Management Systems}\label{subsec:rel_work_collection}

Works intending to collect, manage and preprocess data to be used during the development, test and simulation of algorithms for automated driving applications in general can be divided into two main groups. The approaches of the first group focus on data collection from an external point of view. Therefore, camera systems or  statically positioned drones looking from a birds-eye view onto the scene are used. Whereas the NGSIM data set \cite{colyar2007us} has been very popular, most researchers recently switched over to the more exact and larger highD and inD data sets \cite{krajewski2018highd, bock2019ind}. 

By contrast, the second group relies on dedicated measurement vehicles \cite{ramanishka2018toward, klitzke2019real, chang2019argoverse}. Therefore, the data collection is performed from a moving point of view within the scene. Thus, challenges as occluded vehicles can occur and the data collection mechanism itself can influence the collected behaviors. In exchange, measurement durations for single vehicles can be significantly increased compared to approaches in the first group. 

In addition to the data sets mentioned above, which represent traffic scenes and their objects through numerical descriptions of the traffic scenes and their objects, several optical data sets exist. As examples the popular KITTI \cite{geiger2013vision} and Cityscapes \cite{cordts2016cityscapes} data sets or the upcoming Baidu driving \cite{yu2017baidu} and Audi autonomous driving \cite{geyer2020a2d2} data sets can be mentioned. While such optical data sets are more suitable, for example, for the development of object detection and semantic segmentation algorithms, numerical ones are significantly more useful for object motion predictions.

Additionally, this work focuses on the development of a fleet learning architecture. Consequently, it is desirable to minimize the size of the data to be transmitted. Accordingly, optical data sets are only of secondary interest for the presented article, as images or even videos are much larger compared to sparse numerical object representations. This also applies to works focusing on the collection of pedestrian motion data, such as the popular UCY, ETH, and stanford drone data sets \cite{lerner2007crowds, pellegrini2009you, robicquet2016learning}, as our research focuses the prediction of vehicle motions.

\subsection{Behavior Prediction Approaches}\label{subsec:rel_work_prediction}

According to Lef{\`e}vre \cite{lefevre2014}, behavior prediction approaches can be divided into the three categories: physics-based, maneuver-based, and interaction-aware prediction approaches. \textit{Physics-based approaches} assume that future vehicle motions solely depend on the laws of physics and can be described with simple models such as constant velocity or constant acceleration. A good overview on corresponding approaches is provided in \cite{schubert2008comparison}. By contrast, \textit{maneuver-based approaches} (e.\,g. \cite{wirthmueller2020, schlechtriemen2015will, wirthmueller2019,  wissing2018trajectory, schlechtriemen2014lane, cui2019multimodal, benterki2020artificial}) try to infer the maneuver a driver intends to perform. Finally, \textit{interaction-aware approaches} \cite{lenz2017, bahram2016, zhao2019multi, khakzar2020dual} provide the most advanced motion models by predicting the motions of all vehicles in a given situation simultaneously. In particular, these models consider that all vehicles mutually influence each other.

\cite{wirthmueller2019} uses a categorization, which is more oriented towards the representation of the prediction output and also allows categorizing approaches that cannot be uniquely assigned to one of the aforementioned classes. This categorization distinguishes between approaches for maneuver prediction, position prediction, and hybrid approaches. While maneuver prediction approaches (e.\,g. \cite{schlechtriemen2014lane, bahram2016}) try to infer which one of a fixed set of maneuvers a vehicle will perform, position prediction approaches (e.\,g. \cite{cui2019multimodal, lenz2017, zhao2019multi, khakzar2020dual, fang2020tpnet, altche}) try to infer at which exact position a vehicle will be at a certain future point in time, i.\,e. the latter approaches operate in a continuous space. Finally, hybrid approaches (e.\,g. \cite{wirthmueller2020, schlechtriemen2015will, wirthmueller2019, wissing2018trajectory, benterki2020artificial}) integrate the outputs of maneuver and position prediction approaches into a single or combined model.



\subsection{Contribution}\label{subsec:rel_work_discussion}

As the literature overview has revealed, a lot of research has been spent on data collection for automated driving as well as on motion prediction. However, the presented works presume a setting, where a data set is collected once and afterwards utilized to train and validate prediction models. This procedure, obviously limits the variance as well as the size of the data set. In general, it is not possible to collect a data set covering all relevant corner cases through a single data collection campaign.

As an exception, \cite{vasquez2009growing} uses growing hidden markov models to learn trajectory prediction models for pedestrians and vehicles at a fixed location. Essentially, the drivable space is represented as a discretized graph with edges and nodes, which is updated over runtime. Although this work is in line with our research direction, it cannot be integrated into a moving vehicle and a therefore changing surrounding.

In order to bridge the described research gap, this article contributes in three respects:

\begin{enumerate}[label=C.\arabic*]
\item A fleet learning-based architecture enabling enhanced behavior predictions, especially during challenging external conditions, is presented.
\item As a key part of the described architecture, a prediction watchdog is developed.
\item The necessary modules of the architecture are prototypically implemented and integrated into a testing vehicle, demonstrating the fundamental feasibility of the approach.
\end{enumerate}

\begin{figure*}[ht!]
\centering\includegraphics[width=0.92\textwidth]{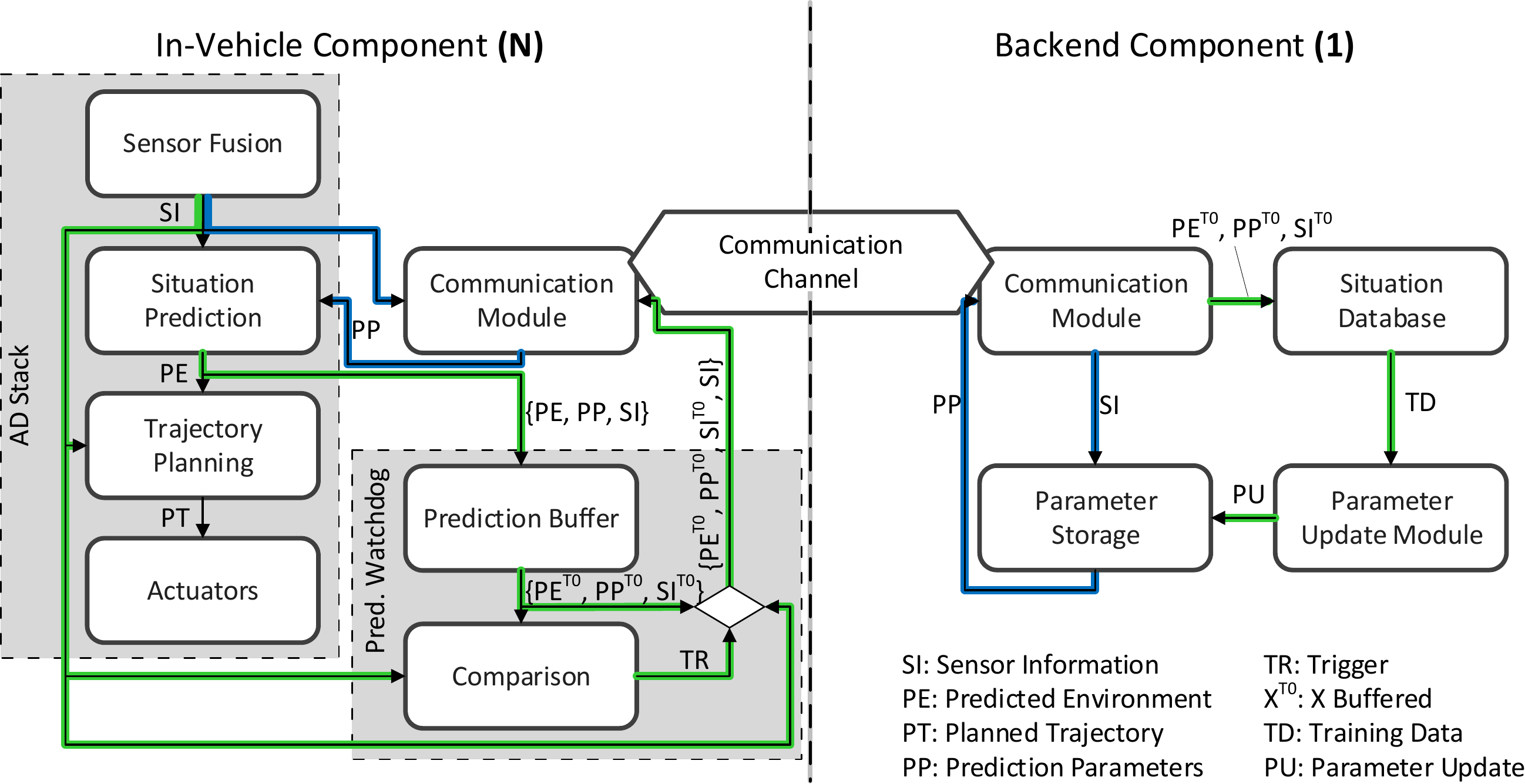}\caption{Overview of the proposed fleet learning architecture. The colored connections highlight the two main loops within the architecture: \newline blue: condition-adaptive parameter request loop; green: data collection and parameter update loop. As indicated by the 1-to-N relation, there are a lot of vehicles with the described in-vehicle component communicating with a single backend component.}\label{fig:architecture}
\end{figure*}

\section{Architecture Concept}\label{sec:arch}

We aim to develop an architecture concept enabling continuous performance improvements to motion prediction systems within a fleet of vehicles. The architecture needs to fulfill the following requirements:

\begin{enumerate}[label=R.\arabic*]
\item \label{req:1} The prediction performance shall be equal over all vehicles of the fleet in any situation. 
\item \label{req:2} Data transmission (e.\,g. via mobile communication) if necessary shall be restricted to a minimum in order to reduce communication costs. 
\item \label{req:3} The prediction module shall produce reliable results even if the system is offline (i.\,.e. if there is no mobile communication signal). 
\item \label{req:4} The overall prediction performance shall increase over lifetime.
\item \label{req:5} All updates to be deployed to the vehicle fleet need to go through a release process.
\end{enumerate}

The developed architecture meeting these requirements is depicted in \autoref{fig:architecture}. From a high-level perspective, the architecture comprises of a communication channel as well as an in-vehicle and a backend component. The in-vehicle component, in turn, comprises seven modules:
\begin{itemize}
\item A sensor fusion module aggregating the raw information from different sensors and providing a consistent representation of the surrounding to other modules.
\item A situation prediction module providing the trajectory planning module with information about the evolution of the current traffic situation. The module's output can be optimized through remote parametrization.
\item A trajectory planning module planning trajectories for the ego-vehicle based on the current sensor information as well as the situation predictions. Good trajectories are characterized by safety and comfort for the passengers. To enable the planning module to generate such trajectories, both inputs need to be as accurate as possible in any situation.
\item Actuators realizing planned trajectories.
\item A prediction buffer storing the current prediction output, the prediction parameters, and the current sensor information until reaching the prediction time.
\item A comparison module comparing the actual positions of the surrounding vehicles with the predictions made some moments ago. If a predicted position differs too much from the actual one, this module triggers the communication module to send a new package of training data containing the buffered prediction (output), the buffered sensor information (input), and the actual position (desired output) to the backend. This mechanism ensures that exactly those situations are detected and used to increase the prediction performance, which are currently handled sub-optimally during the predictions. This contributes to meet requirement \ref{req:2}.
\item A communication module requesting condition-specific prediction parameters from the backend and providing them to the prediction module. This communication module also transmits data backwards over the communication channel.
\end{itemize}


The backend part on the other hand consists of only four modules:
\begin{itemize}
\item A communication module receiving data from the vehicle fleet and transmitting prediction parameters backwards.
\item A condition-adaptive parameter storage, holding the parameters currently used during different situations. Due to the use of a single shared parameter storage for all vehicles of the fleet, requirement \ref{req:1} is met.
\item A situation database storing all data that are necessary to (re-)train a machine learning-based prediction module. In detail:
\begin{itemize}
\item All inputs necessary for the prediction module.
\item The desired output of the prediction module.
\item The external conditions of the measurement (e.\,g. weather or speedlimit).
\end{itemize}
\item A parameter update module using the measurements stored in the situation database to calculate improved parameter values. Before pushing an update to the parameter storage, it is checked whether it increases the performance in all known situations. Then the updated parameters are released resulting in the fulfillment of requirements \ref{req:4} and \ref{req:5}.
\end{itemize}

As further shown in \autoref{fig:architecture}, essentially there are two data loops. The loop emphasized in green collects data that shall enable determining improved prediction parameters in the backend. Within the blue-colored loop, vehicles request condition-adaptive prediction parameters and use them to ensure reliable predictions during all situations. To ensure that the prediction also works in scenarios in which no communication with the backend is possible, the system may fall back to a basic parameter set (fulfilling requirement \ref{req:3}). The latter is initially used as well. To bridge short offline-phases, it is advisable that the vehicles request parameters already in advance if possible. This becomes possible, for example, if the route ahead is known or if it is foreseeable that it will start to rain soon. From the viewpoint of functional safety, it might also be advantageous to rely on a fixed neural network architecture and to solely adjust the weights during parameter updates. This though is also transferable to other prediction techniques.

\begin{figure}[ht!]
\centering\includegraphics[width=0.48\textwidth]{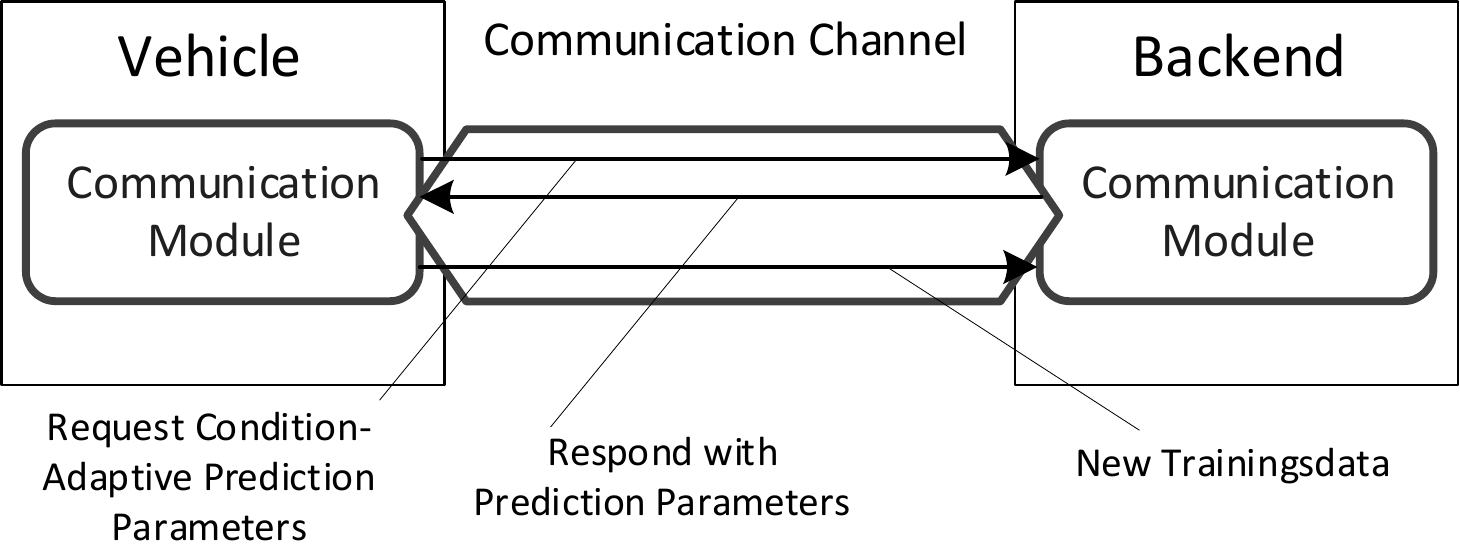}\caption{Overview of the communication channel and the three transmitted message types.}\label{fig:communication}
\end{figure}

\autoref{fig:communication} shows the communication channel and the in-vehicle- and backend-sided communication modules in more detail. Basically, there are three types of messages to be transmitted:

\begin{itemize}
\item New measurements collected by a vehicle that need to be added to the situation database.
\item A request for a parameter set fitting the external conditions a vehicle is faced with.
\item A reply to a parameter request.
\end{itemize}

\section{Prediction Watchdog}\label{sec:mem}

The prediction watchdog as depicted in \autoref{fig:architecture} consists of the prediction buffer and the comparison module. As introduced briefly, it is able to memorize predicted positions as well as the input features that led to the respective model output. Moreover, it compares the actual prediction outputs with the desired one, and triggers the transmission of additional data to the backend. \autoref{ssec:memory-prediction} provides further details on the concept of buffering predictions, whereas \autoref{ssec:comp} outlines the working principle of the comparison module and the triggering.

\subsection{Prediction Buffer}\label{ssec:memory-prediction}
The prediction buffer's function is to hold positions $[\hat{x}_{t_h}, \hat{y}_{t_h}]^T$ predicted at the current point in time $t=t_0$ until reaching the prediction horizon $h$ at $t=t_0+t_h$. In case of an ideal prediction for any given point in time, the memorized point lies on the continuously updated trajectory until the vehicle arrives at that position.


Due a lacking world-fixed coordinate frame, the predicted position has to be fixed to the current environment by changing its vehicle-relative coordinates. The prediction is memorized in lane or also called Frenet coordinates \cite{thorvaldsson2015}, which enable a robust and reasonably precise way of updating the numeric values using Euler integration. The velocity of the ego-vehicle is measured and split into longitudinal and lateral components given the current driving lane, denoted by $\vec{v}=[v_{x}, v_{y}]^T$. For each memory entry, there is a countdown variable $t_c$ that is initialized with the prediction horizon $t_h$ when saving a new prediction. The memorized prediction is updated according to \autoref{eq:mem-eulerint} and \autoref{eq:countdown}:

\begin{equation}
\begin{bmatrix}
\hat{x}_{t_h} \\ \hat{y}_{t_h}
\end{bmatrix}  \leftarrow
\begin{bmatrix}
\hat{x}_{t_h} \\ \hat{y}_{t_h}
\end{bmatrix}- 
\begin{bmatrix}
v_x \\ v_y
\end{bmatrix} \cdot \Delta t
\label{eq:mem-eulerint}
\end{equation}

\begin{equation}
t_c \leftarrow t_c - \Delta t
\label{eq:countdown}
\end{equation}

\begin{figure}[t!]
\centering\includegraphics[width=0.48\textwidth]{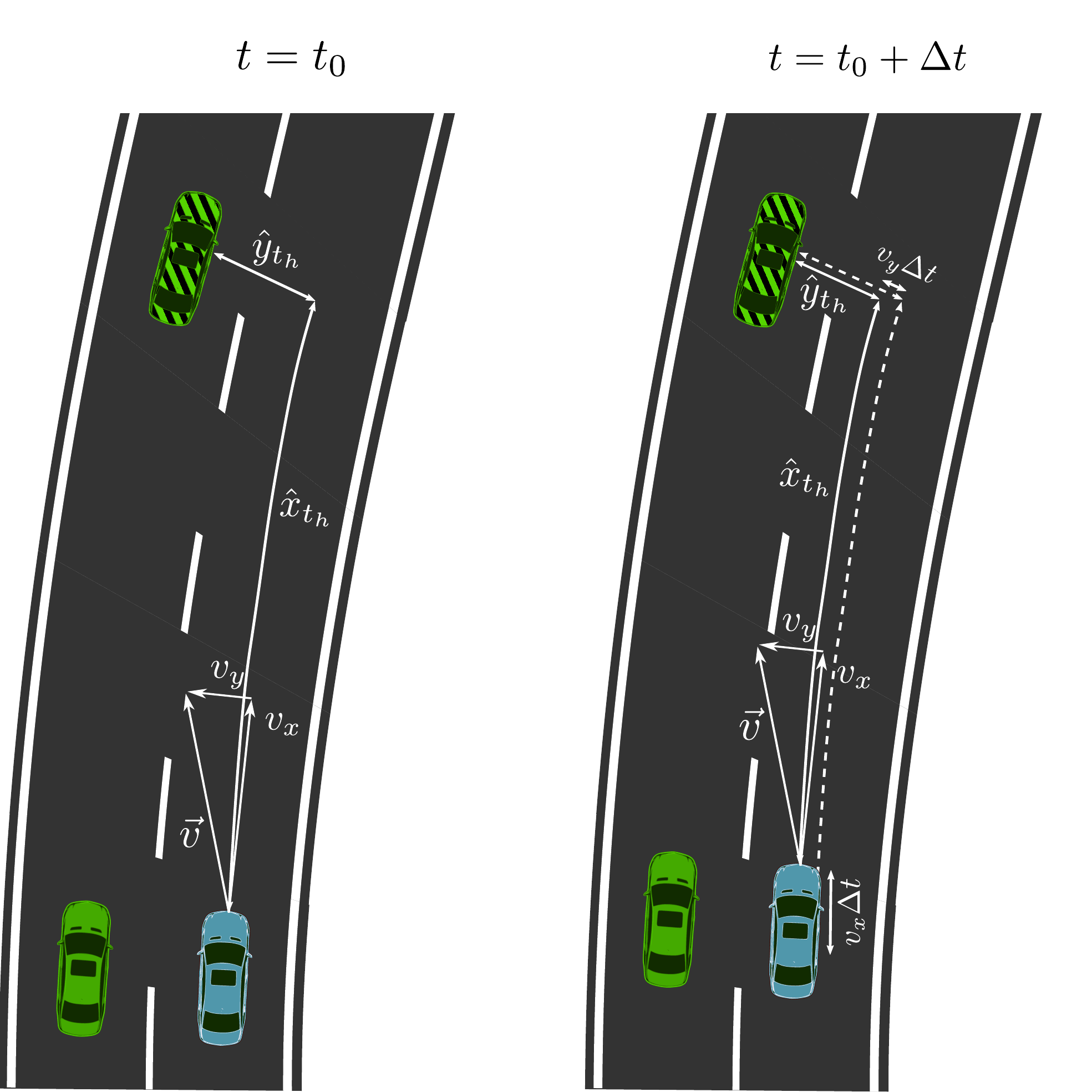}\caption{Illustration of the memory module update using ego-relative coordinates. The striped green vehicle depicts the predicted position of the green vehicle. The vehicle shown in blue is the ego-vehicle.}\label{fig:update}
\end{figure}

 $\Delta t$ corresponds to the time passed since the last model update. \autoref{fig:update} illustrates the process of updating the model. 

Simply put, the predicted position declared relatively to the ego-vehicle performing the prediction is adjusted with the ego-vehicle's movement in each step. As soon as $t_c$ reaches zero, the memorized prediction is not updated anymore and the comparison with the current position can be carried out. Due to the limited model update frequency, the exact moment when the countdown vanishes cannot be captured. A significantly large movement of the vehicle can be caused until the assessment is issued, as the longitudinal velocity can be reasonably high. To deal with that effect a constant-velocity correction in longitudinal direction is performed. This prevents the slack due to finite update frequency to have an effect on the accuracy evaluation. A perfect prediction would feature a residual position of zero in the exact moment the countdown reaches zero.

\begin{figure}[t!]
\centering\includegraphics[height=6.0cm]{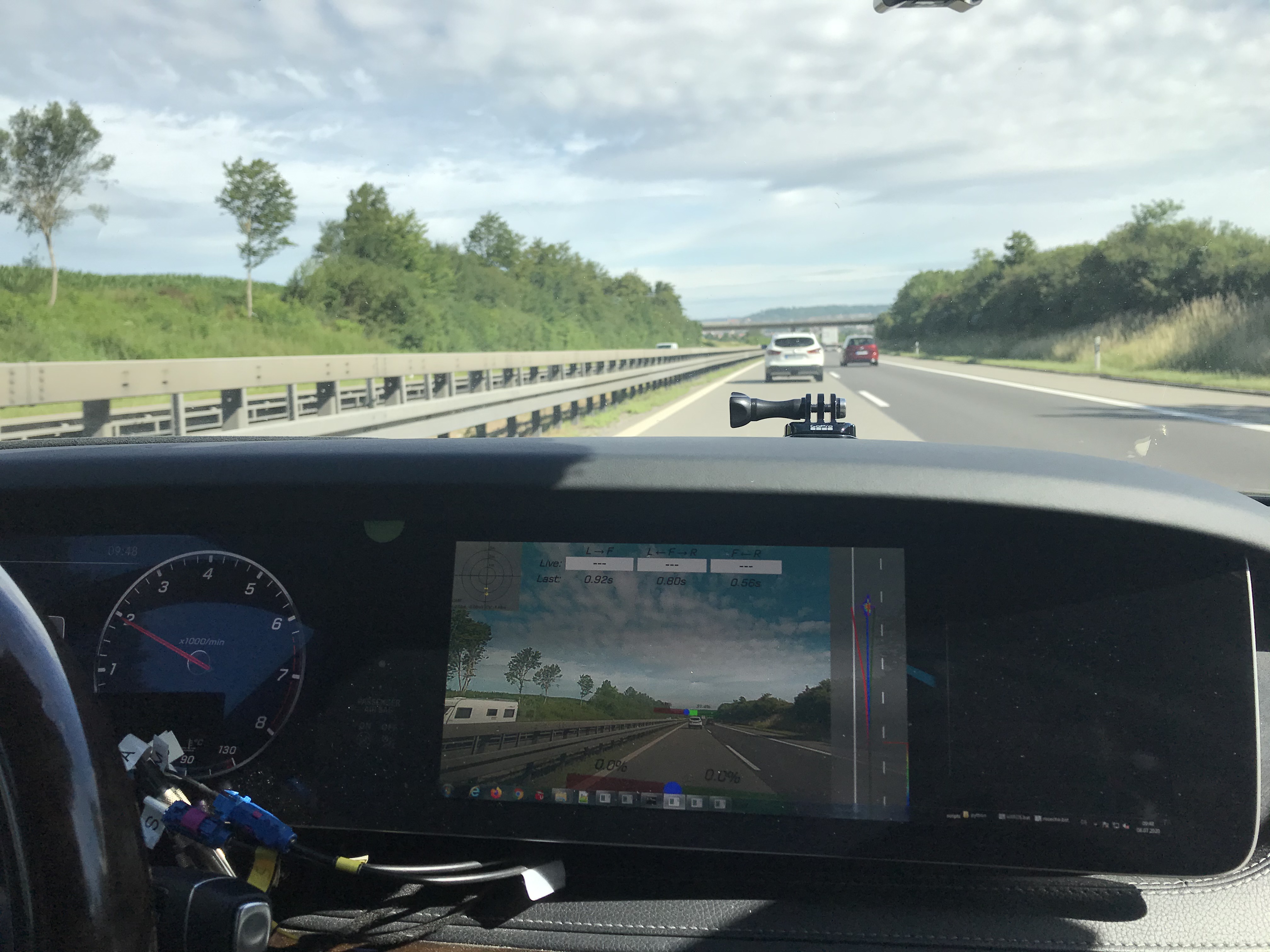}\caption{Image showing the prediction and logging approach integrated into a testing vehicle and connected to an AR visualization.}\label{fig:AR}
\end{figure}

\subsection{Comparison}\label{ssec:comp}

The working principle of the comparison module is rather simple. In order to send the appropriate data to enhance the prediction modules in the backend, it is advisable to select those predictions that are too far away from the desired output, i.\,e. the actual position $[x_{t_h}, y_{t_h}]^T$. Therefore, the comparison module calculates the longitudinal $e_{x, t_h}$ and lateral prediction errors $e_{y, t_h}$ according to \autoref{eq:long_err} and \autoref{eq:lat_err}:

\begin{equation}
e_{x, t_h} = \mid x_{t_h}-\hat{x}_{t_h}\mid
\label{eq:long_err}
\end{equation}

\begin{equation}
e_{y, t_h} = \mid y_{t_h}-\hat{y}_{t_h}\mid
\label{eq:lat_err}
\end{equation}

A a new data package is sent to the backend when exceeding one of the given thresholds $\Theta_x$ and $\Theta_y$ for the two directions.

\section{Application in a Testing Vehicle}\label{sec:application}

In order to demonstrate the fundamental feasibility and benefits of the presented approach, we implemented the described modules in a testing vehicle. To do so, we relied on the findings we published in \cite{wirthmueller2019} with regard to the prediction component as well as on the prediction watchdog presented in \autoref{sec:mem}. Initially, we are restricting ourselves to the prediction of the behavior of the ego-vehicle, as this setting is easier and faster to realize. In general, the same mechanisms can be transfered to surrounding vehicles as well. Our investigations focus on the fundamental feasibility of the data collection loop as well as the quality of the collected data in order to enable model improvements. The design of the communication channel and training adapted prediction models are out of scope of this work. Instead, the data, which would be transferred from the vehicle to the backend in the final application is simply logged in CSV files. This allows for a downstreamed inspection by examining e.\,g. the histograms of the residuals.


The testing vehicle is equipped with a series-like sensor setup consisting of automotive radars and cameras facing the front and the back of the vehicle. Moreover, the testing vehicle is equipped with an additional computing unit, where a ROS environment \cite{quigley2009ros} is deployed. Within the ROS environment, the sensor signals published over the Flexray bus are accessible, allowing for the easy implementation of new functional blocks. The ROS environment is connected to an integrated augmented reality display that allows visualizing the prediction outputs in real time. \autoref{fig:AR} shows an example of the AR visualization of the predictions. A video showing the output for a short sequence can be found on ReserachGate\footnote{\href{https://www.researchgate.net/publication/343392786_Prediction_and_Memory_Module_within_AR_display}{https://www.researchgate.net/publication/343392786\_Prediction\_and \_Memory\_Module\_within\_AR\_display}}. The visualization solution allows for additional visual inspection of the system performance.

In our experiments, the denoted system frequency is 25\,Hz and the threshold for triggering an erroneous lateral prediction logging is set to $\Theta_y =0.2$\,m\footnote{For the sake of simplicity we restricted ourselves to solely investigate the lateral direction here.}. Therefore, the logged data have to show up significantly higher prediction errors than the system shows during normal operation. According to \cite{wirthmueller2019}, the median lateral error should be around 0.11\,m for a prediction horizon of 3\,s. These values cannot be reached in the given experiment, as the prediction models could probably be subject to transfer errors. This is because a different, even though similar, vehicle and sensor setup were used compared to the original work. However, as an approximate estimate the values obtained are sufficient.

\begin{figure}[t!]
\centering\includegraphics[height=6.0cm]{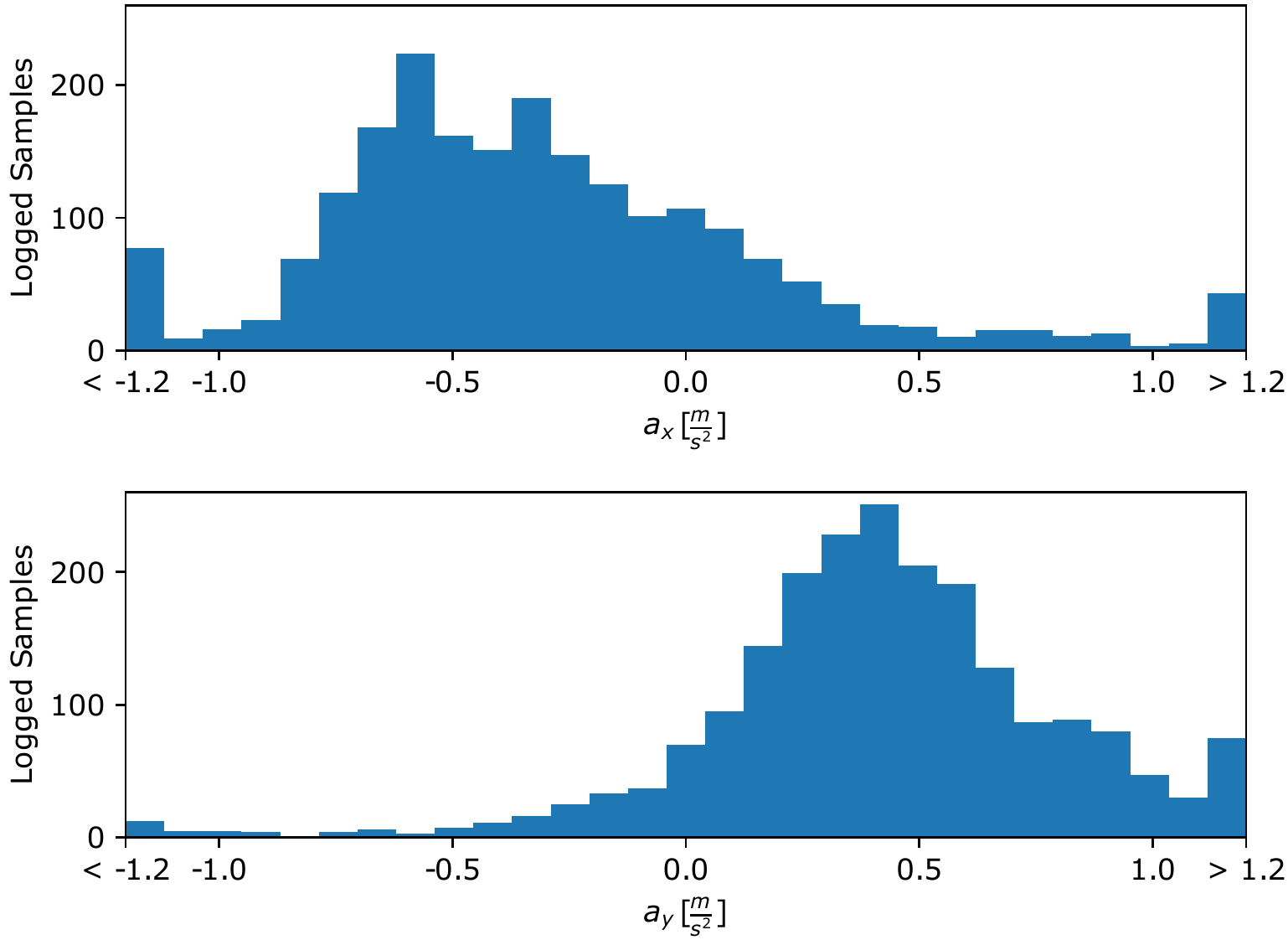}\caption{Histograms showing the distribution of the logged data's longitudinal and lateral acceleration.}\label{fig:eval}
\end{figure}

\autoref{fig:eval} depicts the distributions of the logged samples over the longitudinal $a_x$ (upper part) and lateral $a_y$ acceleration (lower part). This shows the samples collected during several highway test drives with an overall measuring period of more than one hour. According to the visualization, the employed position prediction approach seems to produce more faulty predictions when negative longitudinal or positive lateral accelerations occur. Even this small example with a very limited amount of collected data shows that the presented collection strategy has great potential to enhance training data sets for motion predictions with meaningful samples. Although this example does not refer to external conditions, the same effects could be observed for such, when collecting more data e.\,g. through a fleet of several vehicles.

\section{Summary and Outlook}\label{sec:conclusion}

We presented a new fleet learning-based data collection architecture that ensures continuous improvements of motion predictions of surrounding traffic participants. Especially,  this is beneficial during challenging external conditions. In addition, the relevant elements of the pipeline were prototypically applied to a testing vehicle. Empirical evaluations conducted with the testing vehicle prove the fundamental feasibility of the system. Besides, the investigations show that meaningful samples, which can be used to improve the motion predictions, can be collected.

As the next steps of our research, we will expand the memory component and conduct investigations based on vehicles other than the ego-vehicle. Additionally, we plan a rollout of the data collection architecture on a larger fleet of testing vehicles. Furthermore, we will study the actual improvements of the motion prediction modules being enabled by the collected data with respect to external conditions, as soon as a larger data basis becomes available.



\addtolength{\textheight}{-19.8cm}   

\bibliographystyle{ieeetr}

\bibliography{bib_civts2020}

\end{document}